%% file: irfunduset-main.tex
\documentclass{article}

\usepackage{arxiv}

\usepackage[utf8]{inputenc} 
\usepackage[T1]{fontenc}    
\usepackage{hyperref}       
\usepackage{url}            
\usepackage{booktabs}       
\usepackage{amsfonts}       
\usepackage{nicefrac}       
\usepackage{microtype}      
\usepackage{cleveref}       
\usepackage{lipsum}         
\usepackage{graphicx}
\usepackage[square,numbers,super,sort&compress]{natbib}
\usepackage{doi}

\input{00-commonz}

\title{IRFundusSet: An integrated retinal fundus dataset with a harmonized healthy label}

\date{}

\newif\ifuniqueAffiliation

\ifuniqueAffiliation 
\author{ \href{https://orcid.org/0000-0000-0000-0000}{\includegraphics[scale=0.06]{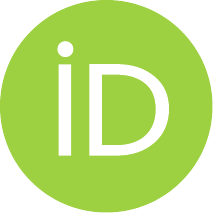}\hspace{1mm}David S.~Hippocampus}\thanks{Use footnote for providing further
		information about author (webpage, alternative
		address)---\emph{not} for acknowledging funding agencies.} \\
	Department of Computer Science\\
	Cranberry-Lemon University\\
	Pittsburgh, PA 15213 \\
	\texttt{hippo@cs.cranberry-lemon.edu} \\
	\And
	\href{https://orcid.org/0000-0000-0000-0000}{\includegraphics[scale=0.06]{orcid.pdf}\hspace{1mm}Elias D.~Striatum} \\
	Department of Electrical Engineering\\
	Mount-Sheikh University\\
	Santa Narimana, Levand \\
	\texttt{stariate@ee.mount-sheikh.edu} \\
}
\else
\usepackage{authblk}

\newbox{\orcid}\sbox{\orcid}{\includegraphics[scale=0.06]{orcid.pdf}} 
\author[1]{%
	\href{https://orcid.org/0009-0000-2080-4979}{\usebox{\orcid}\hspace{1mm}P.~Bilha Githinji}%
}
\author[2]{%
	\href{https://orcid.org/0000-0002-0682-2780}{\usebox{\orcid}\hspace{1mm}Keming Zhao}%
}
\author[2]{%
	\href{https://orcid.org/0000-0000-0000-0000}{\usebox{\orcid}\hspace{1mm}Jiantao Wang}%
}
\author[1]{%
	\href{https://orcid.org/0000-0001-7336-7848}{\usebox{\orcid}\hspace{1mm}Peiwu Qin}%
}
\affil[1]{Tsinghua University, Tsinghua-Berkeley Shenzhen Institute, Shenzen, China}
\affil[2]{Shenzhen Eye Hospital, Jinan University, Shenzhen Eye Institute, Shenzhen, China}
\fi

\renewcommand{\headeright}{} 
\renewcommand{\undertitle}{} 
\renewcommand{\shorttitle}{IRFundusSet: An integrated retinal fundus dataset with a harmonized healthy label}

\hypersetup{
pdftitle={A template for the arxiv style},
pdfsubject={q-bio.NC, q-bio.QM},
pdfauthor={David S.~Hippocampus, Elias D.~Striatum},
pdfkeywords={First keyword, Second keyword, More},
}

\begin{document}
\maketitle

\begin{abstract}	
	\input{\CHAPTERSDIR/1abstract}
\end{abstract}


\section*{Background \& Summary}
\label{sec:dpintro}
\input{\CHAPTERSDIR/2intro}

\section*{Methods}
\label{sec:dpmethod}
\input{\CHAPTERSDIR/3methods}

\section*{Data Records}
\label{sec:dpdata}
\input{\CHAPTERSDIR/4datadesc}

\section*{Technical Validation}
\label{sec:dptech}
\input{\CHAPTERSDIR/5techvalid}

\section*{Usage Notes}
\label{sec:dpuse}
\input{\CHAPTERSDIR/6usageconc}

\section*{Code availability}
\label{sec:dpcode}

The \DATASETNAMEFULL~(\DATASETNAME) is publicly available on Github and Zenodo. The \DATASETNAME Python modules are on Github at https://github.com/bilha-analytics/IRFundusSet, while the independent curated catalogue is on Zenodo at https://zenodo.org/records/10617824. A user guide in the form of a Jupyter Notebook is included on Github as well. 


\section*{Acknowledgements} 
We thank the support from the National Natural Science Foundation of China 31970752; 32350410397; Science, Technology, Innovation Commission of Shenzhen Municipality JSGG20200225150707332, JCYJ20220530143014032, WDZC20200820173710001, WDZC20200821150704001; Shenzhen Medical Academy of Research and Translation, D2301002; Shenzhen Bay Laboratory Open Funding, SZBL2020090501004; Department of Chemical Engineering-iBHE special cooperation joint fund project, DCE-iBHE-2022-3; Tsinghua Shenzhen International Graduate School Cross-disciplinary Research and Innovation Fund Research Plan, JC2022009; and Bureau of Planning, Land and Resources of Shenzhen Municipality (2022) 207.

{
	\bibliographystyle{unsrtnat} 
	\bibliography{\dpaperbib}
}

\end{document}

%% file: 00-commonz.tex
\newcommand{\FIGURESDIR}{.} 

\newcommand{\dpaperbib}{tf-zdpaper}
\newcommand{\dpaperfigz}{\FIGURESDIR}

\newcommand*{\zsubsection}{\subsection*}

\newcommand{\DATASETNAMEFULL}{Integrated Retinal Fundus Set}
\newcommand{\DATASETNAME}{IRFundusSet}
\newcommand{\NORMAL}{\textit{healthy}}

\newcommand{\COHORT}{cohort}

\newcommand{\FUNDUSIMAGEFULL}{retinal fundus color photograph}
\newcommand{\FUNDUSIMG}{RFP}

\newcommand{\NCOHORTS}{10\space}
\newcommand{\NALL}{46064\space}
\newcommand{\NALLCURATED}{25406\space}           
\newcommand{\NALLCURATEDWITHLABEL}{19871\space}  
\newcommand{\NSRCNORMAL}{29708\space}
\newcommand{\NNORMAL}{3515\space}

\newcommand{\NEyePACS}{35108\space}              
\newcommand{\NEyePACSNORM}{25802\space}          
\newcommand{\NEyePACSNORMPERC}{73.5\%\space}     
\newcommand{\NEyePACSCOVER}{14450\space}         
\newcommand{\NEyePACSCOVERPERC}{41.2\%\space}    %

\usepackage{booktabs, threeparttable} 

\crefname{section}{Sec.}{Secs.}
\Crefname{section}{Section}{Sections}
\Crefname{table}{Table}{Tables}
\crefname{table}{Tab.}{Tabs.}

%% file: 1abstract.tex

Ocular conditions are a global concern and computational tools utilizing retinal fundus color photographs can aid in routine screening and management. Obtaining comprehensive and sufficiently sized datasets, however, is non-trivial for the intricate retinal fundus, which exhibits heterogeneities within pathologies, in addition to variations from demographics and acquisition. Moreover, retinal fundus datasets in the public space suffer fragmentation in the organization of data and definition of a healthy observation. 
We present \DATASETNAMEFULL~(\DATASETNAME), a dataset that consolidates, harmonizes and curates several public datasets, facilitating their consumption as a unified whole and with a consistent \emph{is\_normal} label. \DATASETNAME~comprises a Python package that automates harmonization and avails a dataset object in line with the PyTorch approach. Moreover, images are physically reviewed and a new \emph{is\_normal} label is annotated for a consistent definition of a healthy observation. Ten public datasets are initially considered with a total of \NALL images, of which \NALLCURATED are curated for a new \emph{is\_normal} label and \NNORMAL are deemed healthy across the sources. 

%% file: 2intro.tex
Vision impairment and vision-threatening conditions are a growing global concern, with a World Health Organization (WHO) estimate of about 2.2 billion people globally being affected~\cite{world_health_organization_world_2019}. Moreover, the resulting health and economic burden, on both individuals and societies, is expected to worsen with population growth, aging demographics and lifestyle changes~\cite{das_recently_2021,li_cost-effectiveness_2022,world_health_organization_world_2019}. However, the gradual onset of these conditions allows for timely intervention and effective management through routine screening and monitoring~\cite{li_cost-effectiveness_2022,quinn_clinical_2019,grossniklaus_introduction_2015-1, das_recently_2021,zafar_artificial_2022,world_health_organization_world_2019}.

The \FUNDUSIMAGEFULL~(\FUNDUSIMG) is a color image of the posterior region of the retina. It is a widely accepted non-invasive and cost-effective clinical tool for early screening and periodic monitoring of retinal health ~\cite{mishra_fundus_2023,panwar_fundus_2016,khan_global_2021,li_cost-effectiveness_2022}. In addition, computer-aided diagnosis (CAD) tools and artificial intelligence (AI) or machine learning (ML) algorithms utilizing \FUNDUSIMG~images have shown great potential, and are gaining regulatory approvals for restricted cases~\cite{ferro_desideri_upcoming_2022,matta_towards_2023,zafar_artificial_2022,gu_application_2023,iqbal_recent_2022}. 

Despite the progress in CAD and ML solutions, challenges persist in extending advancements to real-world settings, diverse populations and comprehensive disease coverage~\cite{matta_towards_2023,ferro_desideri_upcoming_2022,zafar_artificial_2022,gu_application_2023,iqbal_recent_2022,grzybowski_artificial_2023-2}. Furthermore, a notable strategy in the technologies is a high degree of specialization, utilizing subsets of the retinal features and potentially leaving valuable information unexploited~\cite{iqbal_recent_2022,grzybowski_artificial_2023-2,matta_towards_2023,ferro_desideri_upcoming_2022,zafar_artificial_2022,gu_application_2023}. The retinal fundus is an intricate scene of several anatomical structures, a variety of lesions and multiplicity of relationships between lesions and the diseases that present in the retina~\cite{yannuzzi_retinal_2010-1,grossniklaus_introduction_2015-1,quinn_clinical_2019}. Incorporating such diversity in AI design and modeling might be of additional value.

Creating comprehensive datasets of sufficient scale for medical data is a non-trivial endeavor, requiring expert input and with potential selection biases for reasons such as healthy subjects might not be as inclined to participate. Nonetheless, publicly accessible \FUNDUSIMG~datasets have contributed significantly to advancing research and facilitating clinical translation~\cite{matta_towards_2023,khan_global_2021,gu_application_2023, cen_automatic_2021}. Some of these public datasets capture a myriad of heterogeneities such as variations in image acquisition across centers, different ethnicities and age groups, as well as diverse lesions or pathologies~\cite{matta_towards_2023,khan_global_2021}. For instance, ODIR \cite{noauthor_odir-2019_nodate}, avails 7000 images from several centers and has over 300 pathology labels. Conversely, EyePACS~\cite{noauthor_diabetic_nodate} has over 35000 images from different ethnic groups but with a focus on one disease, Diabetic Retinopathy.

To harness the strengths of various public datasets and attain sufficient scale, however, may detract a research study from its primary objectives. For starters, considerable effort on mundane tasks is required to consume public datasets as a cohesive unit as they have different approaches to how they organize and archive data. Moreover, they adopt different definitions for disease labels including what is a healthy or non-pathological observation, resulting in design choices that might restrict dataset diversity or necessitate additional image curation.

Here we present \DATASETNAMEFULL, a module and a dataset that collates several public datasets to ease their use as a harmonized unit, and curates a common definition of what is a normal or healthy eye. Specific contributions include 
\begin{itemize}
	
	\item {A Python module that consolidates and harmonizes a select collection of ten \FUNDUSIMG~ data sources,  encompassing diversity with regards to collection centers, equipment, ethnicity, age-groups and retinal pathologies. }
	
	\item {Manual review and annotation of images that represent a normal or healthy eye. }
	
	\item {A curated catalog of the entire collection with an additional \emph{is\_normal} label for \NALLCURATEDWITHLABEL images of which \NNORMAL images are determined to be of healthy eyes irrespective of the data source context. } 
	
\end{itemize}

%% file: 3methods.tex

\DATASETNAMEFULL~(\DATASETNAME) is a curated collection of retinal fundus images and their associated pathology labels, which is derived from ten publicly available datasets. In the following subsections, we describe the steps involved in the selection, curation and construction of the final unified dataset. The dataset sources, also referred to as \COHORT s, are summarized in \Cref{tbl:dcohortz}. 

\emph{Ethics statement:} Identified datasets are in the public space and links to their websites or download pages are accordingly referenced. We leave it to the researcher to access and directly download the datasets. 

\zsubsection{Public data sources}
In this version of \DATASETNAME~ a total of \NCOHORTS publicly accessible data sources are considered. The \COHORT s are selected based on ease of access as well as for their potential to represent diversity in retinal fundus images. Together, the selected data sources represent multiple collection centers, several ethnicities and various age groups. Additionally, they include some of the common retinal pathologies like Diabetic Retinopathy (DR), Diabetic Macular Edema (DME), Age-related Macular Degeneration (AMD), Glaucoma, Cataracts, and Pathologic Myopia (PM), encompassing a variety of lesions or pathologies. 

Specifically, the public sources are 
\begin{itemize}
	
	\item{\textit{EyePACS (on Kaggle)\cite{noauthor_data_nodate,noauthor_diabetic_nodate}}: This is a large multi-center and multi-ethnicity dataset collected in the USA that focuses on DR. We retrieve a total of \NEyePACS images of which \NEyePACSNORM are originally labeled as not pathological for DR. We manually cover \NEyePACSCOVER images during curation. This \COHORT~particularly adds diversity with respect to ethnicity, types of cameras and center-specific variations.}
	
	\item{\textit{ODIR \cite{noauthor_odir-2019_nodate, noauthor_ocular_nodate}}: This is a relatively large multi-center dataset from 	Peking University (PKU) International Competition on Ocular Disease Intelligent Recognition (ODIR), China. It covers various retinal conditions including DR, Glaucoma, Cataracts, AMD, Hypertension, PM and some rare conditions. We curate the entire set of images in this dataset. This cohort adds diversity with respect to retinal conditions, cameras and center-specific variations. }
	
	\item{\textit{Kaggle1000 \cite{cen_automatic_2021}}: This dataset entails 39 retinal conditions, including healthy or normal eyes. It is derived from a larger dataset that was used to train an ensemble model for classifying 39 retinal conditions~\cite{cen_automatic_2021}, and avails 1000 fundus images from the Joint Shantou International Eye Center (JSIEC) in China. This \COHORT~brings a diverse set of retinal conditions and lesions to \DATASETNAME. We curate the entire set of images.}
	
	\item{\textit{PAPILA \cite{kovalyk_papila:_2022}}: This is a glaucoma-focused dataset from a single center in Spain. Additionally, it includes images from subjects between the ages of 14 and 90. We curate the entire set of images.}
	
	\item{\textit{IDRiD \cite{porwal_indian_2018}}: The Indian Diabetic Retinopathy Image Dataset (IDRiD) is a single-center dataset from India that looks into DR and DME. This dataset has three task-related subsets - a grading or classification dataset, a segmentation dataset and a dataset for the localization of the optic disc. We focus on the grading dataset where disease labels are included and curate the entire set of those images. It further contributes to the ethnic diversity in \DATASETNAME.}
	
	\item{\textit{CHASEDB1 \cite{fraz_ensemble_2012,fraz_chase_db1_2012}}: This a dataset of school-going children that is also multi-ethnic. It was collected in the UK for research on cardiovascular conditions and is a relatively small dataset. This dataset represents young patients and avails retinal fundus images focused on the optic disc. }
	
	\item{\textit{HRF (high-resolution fundus) \cite{odstrcilik_retinal_2013, budai_robust_2013}}: This data source avails a set of high-resolution images collected from a center in Europe for DR and glaucoma conditions. We curate the entire set of images.}
	
	\item{\textit{FIVE \cite{jin_fives:_2022}}: This a single-center high-resolution dataset that was built for vessel segmentation under healthy and diseased conditions. The retinal diseases covered include DR, AMD and Glaucoma. We curate the entire set of images. }
	
	\item{\textit{STARE \cite{hoover_locating_2000}}: This is one of the long-standing fundus datasets. It covers 14 conditions including healthy eyes. Being one of the older datasets, we hope that it adds diversity to the types of equipment and protocols used. We curate the entire set of images. }
	
	\item{\textit{Retina Cataracts \cite{noauthor_cataract_nodate}}: This is a bit of a wild-card \COHORT~ in that, while publicly available, it is not well documented. We curate the entire set of images.  }
\end{itemize}

\begin{table}[ht!]	
	\centering
	\small\addtolength{\tabcolsep}{-2pt}
	\begin{threeparttable}[c]
		\input{\dpapertblz/tf-describe-cohortz-fini}
	\end{threeparttable}
	\caption{\label{tbl:dcohortz}Data sources or cohorts considered in this version of \DATASETNAME.}
\end{table}

\zsubsection{Consolidating the cohorts} 
\DATASETNAME~ provides a set of Python modules that parse the various directory structures and files to generate a unified catalog of the cohorts. The Python modules contain cohort-specific parsers that map out their directory structures, and extract and collate relevant metadata and condition labels. This is because some datasets do not provide a metadata file or a CSV file of disease labels, instead uniquely encapsulating such information within the directory structure of the dataset. Relatedly, a unique image identifier is created from the relative file path of an image since some datasets employ identical filenames for images belonging to the same subject, relying on the directory structure for context. The result is a unified catalog that seamlessly maps back to the original directory structures, and to which we append additional metadata and the newly curated \emph{is\_normal} label. 

\zsubsection{Curating the normal label}
Extensive literature delves into reviewing retinal fundus images and the phenotypic characteristics of various retinal diseases, offering descriptive notes, classification frameworks, image atlases and comprehensive insights into their manifestation~\cite{quinn_clinical_2019, yannuzzi_retinal_2010-1, grossniklaus_introduction_2015-1, mishra_fundus_2023, shukla_diabetic_2023,yuen_telehealth_2022,muller_ophthalmic_2019, fang_progression_2018, noauthor_pathologic_nodate, kolb_facts_1995, das_recently_2021,masland_neuronal_2012, pm1c, pm22, pm73}. Additionally, these resources also illustrate what healthy eyes look like for different ethnicities, age groups, and equipment in some cases. Furthermore, the guideline~\cite{noauthor_annotation_2021} specifically aims to standardize the use of \FUNDUSIMG~for machine learning purposes. We employ the definitions and guidelines in these works and consider the pathological labels provided by the cohorts to be compliant.

We conduct three rounds of curation with the goal of incrementally refining the quality of the curated labels as we gain more experience through interaction with a large number of images over time. For each curation round, we commence by exploring literature and visual atlases, paying attention to also clarify any concerns we might have from the previous iteration. Secondly, we manually review and curate the \emph{is\_normal} label for each image identified as healthy by the source dataset, eliminating images with conclusive and suspect pathologies. Lastly, we do a final review of only the images that have been identified as healthy in the previous step. The first two steps happen iteratively over consecutive days where they are the primary tasks, while the last step happens after a break of a day or two. The three rounds of data curation happened at least a month apart.

\begin{figure}
	\centering 
	\includegraphics[width=0.65\linewidth]{\dpaperfigz/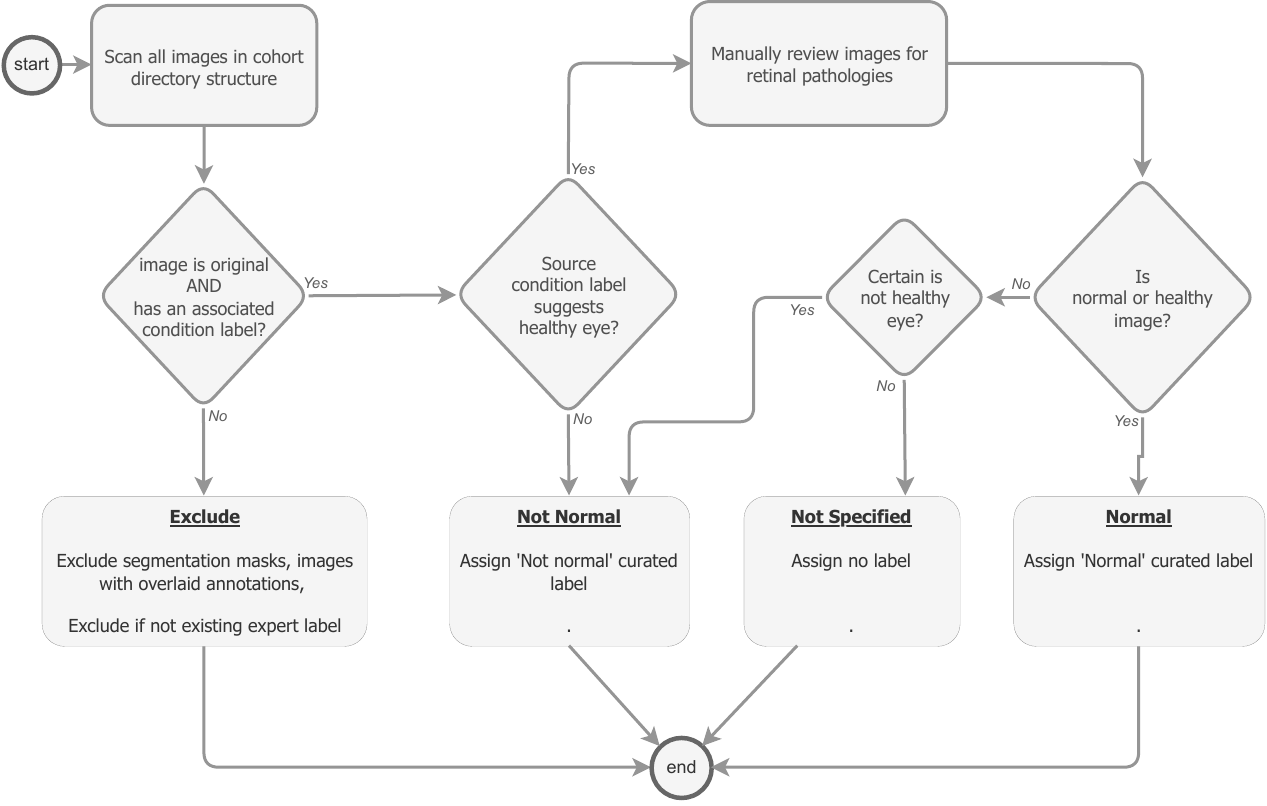}  
	\caption{Flow chart depicting the process for curating \emph{is\_normal} label}
	\label{fig:flow-normz-curate}
\end{figure}

To determine the \emph{is\_normal} label for a given image, we employ the workflow depicted in \Cref{fig:flow-normz-curate}. A primary inclusion requirement is that an image be an original color fundus image for which there is an associated condition label provided by the source dataset. As per the workflow, any image with a source condition label that indicates some pathology gets the label \textit{not normal}. On the other hand, images with a source condition label that suggests a non-pathological eye are subjected to the other steps in the workflow and may not receive the value \textit{normal}. Only images confidently identified as representing healthy eyes are assigned the value \NORMAL~. Moreover, images that are suspected to be pathological but not conclusive are left unlabeled. 

The unified catalog is accordingly updated with the curated \emph{is\_normal} label. Additional metadata is also assigned, including labels indicating whether the image is of a left or right eye, and whether the image is centered around the macular or the optic nerve head. 

\zsubsection{Harmonizing the pixel data} 
Additionally, \DATASETNAME~ includes a set of Python modules that automate the harmonization of the images themselves and provide optional capabilities to statistically standardize the pixel data, thereby aiding in mitigating batch effects. In this step, the images can be updated to have the same image size and file format. Furthermore, statistical harmonization first conducts a within-cohort normalization and then updates the larger unified dataset. Two statistical harmonization methods are included; 1. A standard method that makes use of the mean and standard deviation, and 2. A robust method, which employs the median and inter-quartile range.

%% file: tf-describe-cohortz-fini.tex
\begin{tabular}{lrrccccccccccc}
\toprule
Cohort & n images & n labels & year & region & n centers & FOV & mydriatic & n sizes & n file types \\
\midrule
CHASEDB1 & 28 & 1 & 2011 & UK & 1 & 30 &  & 1 & 1 \\
HRF & 45 & 3 & 2013 & EU & 1 & 45 &  & 1 & 2 \\
STARE & 397 & 42 &  & USA & 2 &  &  & 1 & 1 \\
PAPILA & 488 & 3 & 2018 – 2020 & Spain & 1 & 30 & no & 1 & 1 \\
IDRiD & 516 & 11 & 2009 – 2017 & India & 1 & 50 & yes & 1 & 1 \\
Retina Cataracts & 601 & 4 &  & & &  &  & 1 & 1 \\
FIVE & 800 & 4 & 2016 – 2021 & China & 1 & 50 & yes & 1 & 1 \\
Kaggle1000 & 1000 & 39 & 2009 – 2018 & China & 1 & 35 – 50 & yes & 17 & 3 \\
ODIR & 7000 & 329 &  & China & various & various &  & 101 & 1 \\
EyePACS & 35108 & 5 &  & USA & 7 &  &  & 315 & 1 \\
\bottomrule
\end{tabular}

%% file: 4datadesc.tex

\zsubsection{Python modules}
The Python modules aim to streamline the consolidation process, allowing users to simply download the source datasets and subsequently execute the Python package to parse, catalog and harmonize the retinal fundus images into a unified dataset. The primary functionalities of the Python modules are outlined below, and a user guide in the form of a Jupyter Notebook~\cite{noauthor_jupyter_nodate} is included with the package.

\begin{itemize}
\item{\textit{Generate unified dataset:} This operation generates \DATASETNAME~ and saves it to a specified local directory. It takes as input a list of the cohorts, a desired output directory, the preferred output image size, and the statistical harmonization method if at all. It outputs the harmonized images and a curated catalog in CSV format. This operation may be used from the command line or in batch mode.
}

\item{\textit{Get a dataset:} This operation seamlessly generates or accesses \DATASETNAME~within a ML/AI data pipeline. We define an \DATASETNAME~dataset object as a collection generator that accepts transformations in line with PyTorch~\cite{paszke_pytorch:_2019} style for datasets. For each record in the dataset object, a dictionary is returned, where `image` indexes the retinal fundus image and `target` indexes the curated \emph{is\_normal} label or the source condition label as desired.
} 

\item{\textit{Dataset information:} Additionally, a description of the unified dataset and the cohorts included in it can be obtained.}
\end{itemize}

\zsubsection{Curated catalog}
The curated catalog is a master file that indexes the various images in \DATASETNAME and contains associated metadata and target labels. It is used with the Python package above or may be independently accessed for more involved use cases. The key variables are 

\begin{itemize}
	\item{\textit{image\_id}: This is a unique identifier for each image observation and is derived from a relative file path. }
	\item{\textit{cohort}: This identifies the source dataset or cohort from which an image was obtained }
	\item{\textit{is\_normal}: This is the curated \textit{'is normal'} label, indicating if an observation is of a healthy eye or not. }
	\item{\textit{src\_is\_normal}: This is a \textit{'is normal'} label based on the original cohort-specific label of what is not pathological. }
	\item{\textit{src\_condition}: This is the more granular disease target label as provided by the source dataset. }
	\item{\textit{is\_left\_or\_right\_eye}: This indicates if an eye is left or right eye. }
	\item{\textit{is\_mac\_or\_onh\_centered}: This indicates if the image is centered around the macular or optic nerve head of the retina. }
	\item{\textit{split}: This is the train/test split assignment as per the source dataset.}
\end{itemize}

%% file: 5techvalid.tex

\zsubsection{Data characteristics}
This iteration of \DATASETNAME~consolidates \NCOHORTS public datasets, which entail a total of \NALL images. Among these images, \NSRCNORMAL have a source condition label indicating a non-pathological status within the context of their respective datasets. \NALLCURATED images are subjected to the curation and annotation process, resulting in \NALLCURATEDWITHLABEL images being assigned a new \emph{is\_normal} label. Furthermore, the breakdown of the \NALLCURATED  images with a new \emph{is\_normal} label by size of dataset is 14450 images from EyePACS, 7000 images from ODIR, 1000 images from Kaggle1000, and 2956 images from the remaining seven cohorts. \Cref{tbl:dcohortz-desc} and \Cref{tbl:dnormz} further break down the numbers by cohort and by \emph{is\_normal} label. 

Of the curated images, \NNORMAL images are determined to be of healthy eyes. The proportion of healthy images in the curated dataset is now 14.0\%, a drop from an average of 33.1\%. Excluding the EyePACS dataset due to its inordinate size, the proportion of healthy images in the rest of the curated dataset is 13.9\%, a drop from an average of 28.6\%.

\begin{figure}[ht!]
	\centering
	\includegraphics[width=.4\linewidth]{\dpaperfigz/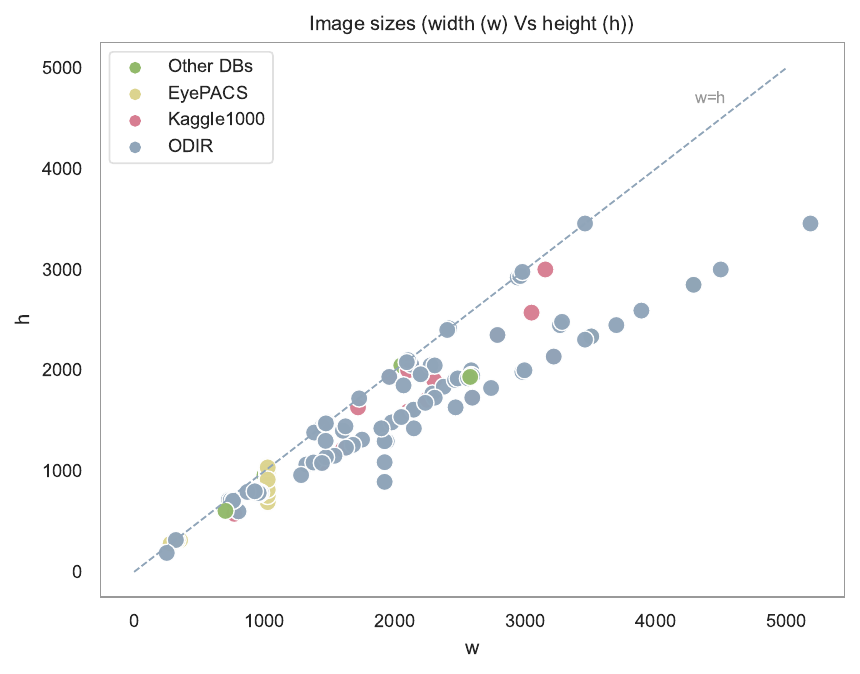} 	
	\includegraphics[width=.4\linewidth]{\dpaperfigz/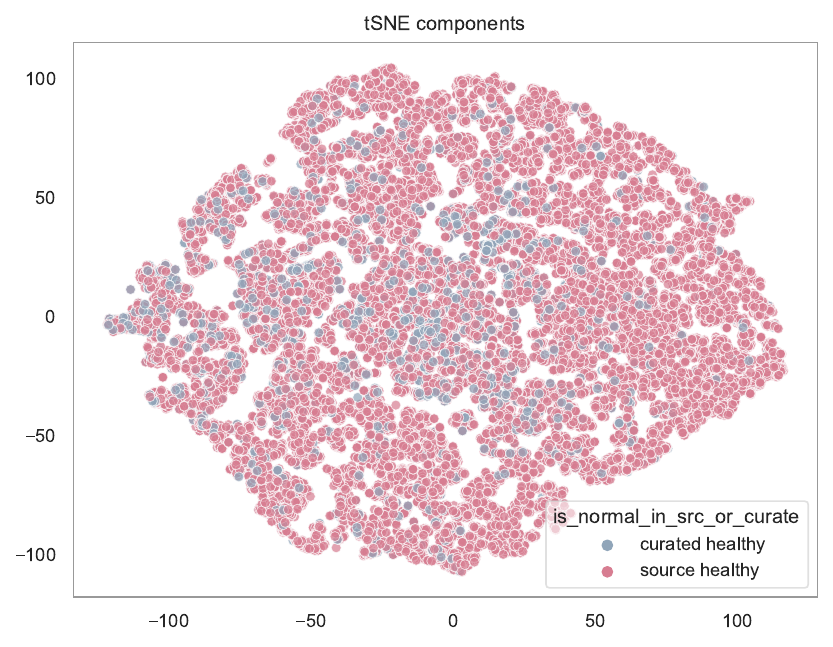} 	
	\caption{The first plot summarizes the aspect ratio of the images. The second plot is a t-SNE visualization of the harmonized dataset exploring for variations that isolate the newly curated \emph{is\_normal}  label }
	\label{fig:dmeta-distz}
\end{figure}

File metadata analysis reveals four main file types and 430 unique image sizes. The four file types are \textit{PNG}, \textit{JPEG}, \textit{TIF} and \textit{PPM}, with variations observed in spelling and case usage. For instance \textit{JPEG} might also be spelled as \textit{JPG}, \textit{jpg} or \textit{jpeg}.  Additionally, different aspect ratios are observed as demonstrated in the first plot in \Cref{fig:dmeta-distz}. The image height is smaller than the width for 92.5\% of the images, and only 2.3\% of the images have a square shape. The smallest image resolution by image height spans 188 pixels and can be found in the ODIR dataset, while the largest image width is 5184 pixels and is also from the ODIR dataset. The average image size in height and width is $(1061 \pm 514, 1289 \pm 675)$ and most of the variation is due to the ODIR dataset. Harmonizing the datasets standardizes the image size and file format. 

The scatter plot in \Cref{fig:dmeta-distz} is a t-SNE visualization of the images after resizing them to $(256, 256)$ and normalizing them using the mean and standard deviation. It qualitatively explores for potential biases accompanying the newly curated \emph{is\_normal} label. 

\begin{table}[ht!]	
	\centering
	\small\addtolength{\tabcolsep}{-2pt}
	\begin{threeparttable}[c]
		\input{\dpapertblz/tf-composition-by-cohorts}
		\begin{tablenotes}
			\item{100\% of images are curated for all cohorts except for EyePACS, where 41\% (14450) images are curated}
			\item{Total values are obtained as sum for `n` columns and average for `\%` columns } 
		\end{tablenotes}
	\end{threeparttable}
	\caption{\label{tbl:dcohortz-desc} Composition by cohorts}
\end{table}

\begin{table}[ht!]		
	\centering
	\small\addtolength{\tabcolsep}{-2pt}
	\begin{threeparttable}[c]
		\input{\dpapertblz/tf-composition-by-is-normal-label}	
	\end{threeparttable}
	\caption{\label{tbl:dnormz} Composition by new \textit{is\_normal} label}
\end{table}

\zsubsection{Notes}
A qualitative review of the pixel data, as depicted in the t-SNE plot, reveals no apparent patterns that are associated with the new \emph{is\_normal} label. In addition, the statics for EyePACS mirror those of other cohorts, suggesting consistent patterns despite curating a subset of EyePACS and EyePACS being inordinately large. Moreover, there is a relatively balanced distribution of left and right eyes in the source datasets and this persists in the curated \DATASETNAME~dataset as well. 

\textbf{Attrition:} Qualitatively, we observed that tessellation, blur and suspect bright lesions were some of the common reasons for reclassifying an image labeled as not pathological by a source dataset. We wonder if myopic eyes may be considered non-pathological in the context of other diseases.

\textbf{EyePACS:} The EyePACS dataset in this iteration is one of the largest public retinal fundus datasets. It also contains one of the largest proportion of images labeled as non-pathological, with a total of \NEyePACSNORM images (\NEyePACSNORMPERC of the dataset) labeled as not pathological for diabetic retinopathy. It is unclear at this stage, however, what proportion of those images is from preprocessing updates in the source dataset. With the time and resources we had, we managed to cover \NEyePACSCOVERPERC of this dataset (\NEyePACSCOVER images) in this version of \DATASETNAME. 

\textbf{Future opportunities:} In addition to harmonizing the pixel data and the \emph{is\_normal} label, the more granular condition labels may also benefit from consolidation. This might entail exploring ophthalmic classification systems and natural language processing tasks to arrive at labels that are not only cohesive but also practical for different tasks.

%% file: tf-composition-by-cohorts.tex
\begin{tabular}{llrcccc}
\toprule
  & n images & n src normal & \% left eye & \% curated & \% old was normal & \% new is normal \\
\midrule
CHASEDB1 & 28 & 0 & 0.50 & 1.00 & 0.000 & 0.000 \\
HRF & 45 & 15 & 0.47 & 1.00 & 0.333 & 0.289 \\
STARE & 397 & 36 & 0.46 & 1.00 & 0.091 & 0.076 \\
PAPILA & 488 & 333 & 0.50 & 1.00 & 0.682 & 0.168 \\
IDRiD & 597 & 168 & 0.52 & 1.00 & 0.281 & 0.201 \\
Retina Cataracts & 601 & 300 & 0.50 & 1.00 & 0.499 & 0.186 \\
FIVE & 800 & 200 & 0.47 & 1.00 & 0.250 & 0.176 \\
Kaggle1000 & 1000 & 38 & 0.42 & 1.00 & 0.038 & 0.038 \\
ODIR & 7000 & 2816 & 0.50 & 1.00 & 0.402 & 0.117 \\
EyePACS & 35108 & 25802 & 0.50 & 0.41 & 0.735 & 0.149 \\
\midrule
Total & 46064 & 29708 & 0.48 & 0.94 & 0.331 & 0.140 \\
Without EyePACS & 10956 & 3906 & 0.48 & 1.00 & 0.286 & 0.139 \\
\bottomrule
\end{tabular}

%% file: tf-composition-by-is-normal-label.tex
\begin{tabular}{lcccc}
\toprule
Curated label & n images & n images curated & \% of all images & \% left eye \\
\midrule
Normal & 3515 & 3515 & 0.08 & 0.49 \\
Not normal & 16356 & 16356 & 0.36 & 0.49 \\
\midrule
Total with label & 19871 & 19871 & 0.43 & 0.49 \\
No label assigned & 26193 & 5535 & 0.57 & 0.50 \\
\bottomrule
\end{tabular}

%% file: 6usageconc.tex

To use the \DATASETNAME~package, a user first downloads the cohort datasets and unzips them on their machine. Secondly, the user specifies which cohorts to include in \DATASETNAME~and where they are saved locally using a template configuration file that is included in the package. Lastly, the user can opt to generate the dataset from the command line or access it within an ML/AI data pipeline. A user guide in the form of a Jupyter Notebook is also available with the dataset modules. 

While we attempt to include as much diversity in this version of the collection, it is not necessarily exhaustive. Additionally, the statistical normalization approaches depend on cohort statistics and, therefore, necessitate a regeneration of the dataset whenever its composition is updated.

Users should cite this paper in any research output that utilizes this dataset in any form and appropriately acknowledge its contribution.

%% file: irfunduset-main.bbl
\begin{thebibliography}{42}
\providecommand{\natexlab}[1]{#1}
\providecommand{\url}[1]{\texttt{#1}}
\expandafter\ifx\csname urlstyle\endcsname\relax
  \providecommand{\doi}[1]{doi: #1}\else
  \providecommand{\doi}{doi: \begingroup \urlstyle{rm}\Url}\fi

\bibitem[{World Health
  Organization}(2019)]{world_health_organization_world_2019}
{World Health Organization}.
\newblock \emph{World report on vision}.
\newblock World Health Organization, Geneva, 2019.
\newblock ISBN 9789241516570.
\newblock URL \url{https://iris.who.int/handle/10665/328717}.

\bibitem[Das et~al.(2021)Das, Takkar, Sivaprasad, Thanksphon, Taylor,
  Wiedemann, Nemeth, Nayar, Rani, and Khandekar]{das_recently_2021}
Taraprasad Das, Brijesh Takkar, Sobha Sivaprasad, Thamarangsi Thanksphon, Hugh
  Taylor, Peter Wiedemann, Janos Nemeth, Patanjali~D. Nayar, Padmaja~Kumari
  Rani, and Rajiv Khandekar.
\newblock Recently updated global diabetic retinopathy screening guidelines:
  commonalities, differences, and future possibilities.
\newblock \emph{Eye}, 35\penalty0 (10):\penalty0 2685--2698, October 2021.
\newblock ISSN 1476-5454.
\newblock \doi{10.1038/s41433-021-01572-4}.
\newblock URL \url{https://www.nature.com/articles/s41433-021-01572-4}.

\bibitem[Li et~al.(2022)Li, Yang, Zhang, Bai, Du, Sun, Tang, Wang, and
  Liu]{li_cost-effectiveness_2022}
Ruyue Li, Ziwei Yang, Yue Zhang, Weiling Bai, Yifan Du, Runzhou Sun, Jianjun
  Tang, Ningli Wang, and Hanruo Liu.
\newblock Cost-effectiveness and cost-utility of traditional and telemedicine
  combined population-based age-related macular degeneration and diabetic
  retinopathy screening in rural and urban {China}.
\newblock \emph{The Lancet Regional Health - Western Pacific}, 23:\penalty0
  100435, June 2022.
\newblock ISSN 26666065.
\newblock \doi{10.1016/j.lanwpc.2022.100435}.
\newblock URL
  \url{https://linkinghub.elsevier.com/retrieve/pii/S2666606522000505}.

\bibitem[Quinn et~al.()Quinn, Csincsik, Flynn, Curcio, Kiss, Sadda, Hogg, Peto,
  and Lengyel]{quinn_clinical_2019}
Nicola Quinn, Lajos Csincsik, Erin Flynn, Christine~A. Curcio, Szilard Kiss,
  {SriniVas}~R. Sadda, Ruth Hogg, Tunde Peto, and Imre Lengyel.
\newblock The clinical relevance of visualising the peripheral retina.
\newblock \emph{Progress in Retinal and Eye Research}, 68:\penalty0 83--109.
\newblock ISSN 13509462.
\newblock \doi{10.1016/j.preteyeres.2018.10.001}.
\newblock URL
  \url{https://linkinghub.elsevier.com/retrieve/pii/S1350946218300399}.

\bibitem[Grossniklaus et~al.()Grossniklaus, Geisert, and
  Nickerson]{grossniklaus_introduction_2015-1}
Hans~E. Grossniklaus, Eldon~E. Geisert, and John~M. Nickerson.
\newblock Introduction to the retina.
\newblock \emph{Progress in Molecular Biology and Translational Science},
  134:\penalty0 383--396.
\newblock ISSN 1878-0814.
\newblock \doi{10.1016/bs.pmbts.2015.06.001}.

\bibitem[Zafar et~al.(2022)Zafar, Mahjoub, Mehta, Domalpally, and
  Channa]{zafar_artificial_2022}
Sidra Zafar, Heba Mahjoub, Nitish Mehta, Amitha Domalpally, and Roomasa Channa.
\newblock Artificial {Intelligence} {Algorithms} in {Diabetic} {Retinopathy}
  {Screening}.
\newblock \emph{Current Diabetes Reports}, 22\penalty0 (6):\penalty0 267--274,
  June 2022.
\newblock ISSN 1539-0829.
\newblock \doi{10.1007/s11892-022-01467-y}.
\newblock URL \url{https://doi.org/10.1007/s11892-022-01467-y}.

\bibitem[Mishra and Tripathy()]{mishra_fundus_2023}
Chitaranjan Mishra and Koushik Tripathy.
\newblock Fundus camera.
\newblock In \emph{{StatPearls}}. {StatPearls} Publishing.
\newblock URL \url{http://www.ncbi.nlm.nih.gov/books/NBK585111/}.

\bibitem[Panwar et~al.()Panwar, Huang, Lee, Keane, Chuan, Richhariya, Teoh,
  Lim, and Agrawal]{panwar_fundus_2016}
Nishtha Panwar, Philemon Huang, Jiaying Lee, Pearse~A. Keane, Tjin~Swee Chuan,
  Ashutosh Richhariya, Stephen Teoh, Tock~Han Lim, and Rupesh Agrawal.
\newblock Fundus photography in the 21st century—a review of recent
  technological advances and their implications for worldwide healthcare.
\newblock \emph{Telemedicine Journal and e-Health}, 22\penalty0 (3):\penalty0
  198--208.
\newblock ISSN 1530-5627.
\newblock \doi{10.1089/tmj.2015.0068}.
\newblock URL \url{https://www.ncbi.nlm.nih.gov/pmc/articles/PMC4790203/}.

\bibitem[Khan et~al.()Khan, Liu, Nath, Korot, Faes, Wagner, Keane, Sebire,
  Burton, and Denniston]{khan_global_2021}
Saad~M Khan, Xiaoxuan Liu, Siddharth Nath, Edward Korot, Livia Faes,
  Siegfried~K Wagner, Pearse~A Keane, Neil~J Sebire, Matthew~J Burton, and
  Alastair~K Denniston.
\newblock A global review of publicly available datasets for ophthalmological
  imaging: barriers to access, usability, and generalisability.
\newblock \emph{The Lancet Digital Health}, 3\penalty0 (1):\penalty0 e51--e66.
\newblock ISSN 25897500.
\newblock \doi{10.1016/S2589-7500(20)30240-5}.
\newblock URL
  \url{https://linkinghub.elsevier.com/retrieve/pii/S2589750020302405}.

\bibitem[Ferro~Desideri et~al.(2022)Ferro~Desideri, Rutigliani, Corazza,
  Nastasi, Roda, Nicolo, Traverso, and Vagge]{ferro_desideri_upcoming_2022}
Lorenzo Ferro~Desideri, Carola Rutigliani, Paolo Corazza, Andrea Nastasi,
  Matilde Roda, Massimo Nicolo, Carlo~Enrico Traverso, and Aldo Vagge.
\newblock The upcoming role of {Artificial} {Intelligence} ({AI}) for retinal
  and glaucomatous diseases.
\newblock \emph{Journal of Optometry}, 15\penalty0 (Suppl 1):\penalty0
  S50--S57, 2022.
\newblock ISSN 1888-4296.
\newblock \doi{10.1016/j.optom.2022.08.001}.
\newblock URL \url{https://www.ncbi.nlm.nih.gov/pmc/articles/PMC9732476/}.

\bibitem[Matta et~al.()Matta, Lamard, Conze, Le~Guilcher, Lecat, Carette,
  Basset, Massin, Rottier, Cochener, and Quellec]{matta_towards_2023}
Sarah Matta, Mathieu Lamard, Pierre-Henri Conze, Alexandre Le~Guilcher,
  Clément Lecat, Romuald Carette, Fabien Basset, Pascale Massin, Jean-Bernard
  Rottier, Béatrice Cochener, and Gwenolé Quellec.
\newblock Towards population-independent, multi-disease detection in fundus
  photographs.
\newblock \emph{Scientific Reports}, 13\penalty0 (1):\penalty0 11493.
\newblock ISSN 2045-2322.
\newblock \doi{10.1038/s41598-023-38610-y}.
\newblock URL \url{https://www.nature.com/articles/s41598-023-38610-y}.

\bibitem[Gu et~al.(2023)Gu, Wang, Jiang, Xu, Wang, Liu, Yuan, Abudureyimu,
  Wang, Lu, Li, Wu, Dong, Chen, Wang, Zhang, Wei, Qiu, Zheng, Liu, and
  Chen]{gu_application_2023}
Chufeng Gu, Yujie Wang, Yan Jiang, Feiping Xu, Shasha Wang, Rui Liu, Wen Yuan,
  Nurbiyimu Abudureyimu, Ying Wang, Yulan Lu, Xiaolong Li, Tao Wu, Li~Dong,
  Yuzhong Chen, Bin Wang, Yuncheng Zhang, Wen~Bin Wei, Qinghua Qiu, Zhi Zheng,
  Deng Liu, and Jili Chen.
\newblock Application of artificial intelligence system for screening multiple
  fundus diseases in {Chinese} primary healthcare settings: a real-world,
  multicentre and cross-sectional study of 4795 cases.
\newblock \emph{British Journal of Ophthalmology}, March 2023.
\newblock ISSN 0007-1161, 1468-2079.
\newblock \doi{10.1136/bjo-2022-322940}.
\newblock URL
  \url{https://bjo.bmj.com/content/early/2023/03/05/bjo-2022-322940}.

\bibitem[Iqbal et~al.()Iqbal, Khan, Naveed, Naqvi, and
  Nawaz]{iqbal_recent_2022}
Shahzaib Iqbal, Tariq~M. Khan, Khuram Naveed, Syed~S. Naqvi, and Syed~Junaid
  Nawaz.
\newblock Recent trends and advances in fundus image analysis: A review.
\newblock \emph{Computers in Biology and Medicine}, 151:\penalty0 106277.
\newblock ISSN 00104825.
\newblock \doi{10.1016/j.compbiomed.2022.106277}.
\newblock URL
  \url{https://linkinghub.elsevier.com/retrieve/pii/S0010482522009854}.

\bibitem[Grzybowski et~al.()Grzybowski, Singhanetr, Nanegrungsunk, and
  Ruamviboonsuk]{grzybowski_artificial_2023-2}
Andrzej Grzybowski, Panisa Singhanetr, Onnisa Nanegrungsunk, and Paisan
  Ruamviboonsuk.
\newblock Artificial intelligence for diabetic retinopathy screening using
  color retinal photographs: From development to deployment.
\newblock \emph{Ophthalmology and Therapy}, 12\penalty0 (3):\penalty0
  1419--1437.
\newblock ISSN 2193-8245.
\newblock \doi{10.1007/s40123-023-00691-3}.

\bibitem[Yannuzzi()]{yannuzzi_retinal_2010-1}
Lawrence~A. Yannuzzi.
\newblock \emph{The retinal atlas: searchable full text online}.
\newblock Elsevier, Saunders.
\newblock ISBN 978-0-7020-3320-9.

\bibitem[Cen et~al.(2021)Cen, Ji, Lin, Ju, Lin, Li, Wang, Yang, Liu, Tan, Tan,
  Li, Wang, Zheng, Xiong, Wu, Jiang, Wu, Huang, Shi, Chen, Yang, Zhang, Luo,
  Huang, Zhang, Huang, Ng, Chen, Chen, Pang, and Zhang]{cen_automatic_2021}
Ling-Ping Cen, Jie Ji, Jian-Wei Lin, Si-Tong Ju, Hong-Jie Lin, Tai-Ping Li, Yun
  Wang, Jian-Feng Yang, Yu-Fen Liu, Shaoying Tan, Li~Tan, Dongjie Li, Yifan
  Wang, Dezhi Zheng, Yongqun Xiong, Hanfu Wu, Jingjing Jiang, Zhenggen Wu,
  Dingguo Huang, Tingkun Shi, Binyao Chen, Jianling Yang, Xiaoling Zhang,
  Li~Luo, Chukai Huang, Guihua Zhang, Yuqiang Huang, Tsz~Kin Ng, Haoyu Chen,
  Weiqi Chen, Chi~Pui Pang, and Mingzhi Zhang.
\newblock Automatic detection of 39 fundus diseases and conditions in retinal
  photographs using deep neural networks.
\newblock \emph{Nature Communications}, 12\penalty0 (1):\penalty0 4828, August
  2021.
\newblock ISSN 2041-1723.
\newblock \doi{10.1038/s41467-021-25138-w}.
\newblock URL \url{https://www.nature.com/articles/s41467-021-25138-w}.

\bibitem[noa({\natexlab{a}})]{noauthor_odir-2019_nodate}
{ODIR}-2019 - {Grand} {Challenge}, {\natexlab{a}}.
\newblock URL \url{https://odir2019.grand-challenge.org/}.

\bibitem[noa({\natexlab{b}})]{noauthor_diabetic_nodate}
Diabetic {Retinopathy} {Detection}, {\natexlab{b}}.
\newblock URL
  \url{https://kaggle.com/competitions/diabetic-retinopathy-detection}.

\bibitem[noa({\natexlab{c}})]{noauthor_data_nodate}
Data {Analysis}, {\natexlab{c}}.
\newblock URL \url{https://www.eyepacs.com/data-analysis}.

\bibitem[noa({\natexlab{d}})]{noauthor_ocular_nodate}
Ocular {Disease} {Recognition}, {\natexlab{d}}.
\newblock URL
  \url{https://www.kaggle.com/datasets/andrewmvd/ocular-disease-recognition-odir5k}.

\bibitem[Kovalyk et~al.(2022)Kovalyk, Morales-Sánchez, Verdú-Monedero,
  Sellés-Navarro, Palazón-Cabanes, and Sancho-Gómez]{kovalyk_papila:_2022}
Oleksandr Kovalyk, Juan Morales-Sánchez, Rafael Verdú-Monedero, Inmaculada
  Sellés-Navarro, Ana Palazón-Cabanes, and José-Luis Sancho-Gómez.
\newblock {PAPILA}: {Dataset} with fundus images and clinical data of both eyes
  of the same patient for glaucoma assessment.
\newblock \emph{Scientific Data}, 9\penalty0 (1):\penalty0 291, June 2022.
\newblock ISSN 2052-4463.
\newblock \doi{10.1038/s41597-022-01388-1}.
\newblock URL \url{https://www.nature.com/articles/s41597-022-01388-1}.

\bibitem[Porwal(2018)]{porwal_indian_2018}
Prasanna Porwal.
\newblock Indian {Diabetic} {Retinopathy} {Image} {Dataset} ({IDRiD}), April
  2018.
\newblock URL
  \url{https://ieee-dataport.org/open-access/indian-diabetic-retinopathy-image-dataset-idrid}.

\bibitem[Fraz et~al.(2012{\natexlab{a}})Fraz, Remagnino, Hoppe, Uyyanonvara,
  Rudnicka, Owen, and Barman]{fraz_ensemble_2012}
Muhammad~Moazam Fraz, Paolo Remagnino, Andreas Hoppe, Bunyarit Uyyanonvara,
  Alicja~R. Rudnicka, Christopher~G. Owen, and Sarah~A. Barman.
\newblock An {Ensemble} {Classification}-{Based} {Approach} {Applied} to
  {Retinal} {Blood} {Vessel} {Segmentation}.
\newblock \emph{IEEE Transactions on Biomedical Engineering}, 59\penalty0
  (9):\penalty0 2538--2548, September 2012{\natexlab{a}}.
\newblock ISSN 1558-2531.
\newblock \doi{10.1109/TBME.2012.2205687}.
\newblock URL \url{https://ieeexplore.ieee.org/document/6224174/}.

\bibitem[Fraz et~al.(2012{\natexlab{b}})Fraz, Remagnino, Hoppe, Uyyanonvara,
  Rudnicka, Owen, and Barman]{fraz_chase_db1_2012}
Muhammad~Moazam Fraz, Paolo Remagnino, Andreas Hoppe, Bunyarit Uyyanonvara,
  Alicja~R. Rudnicka, Christopher~G. Owen, and Sarah~A. Barman.
\newblock {CHASE}\_db1 retinal vessel reference dataset, June
  2012{\natexlab{b}}.
\newblock URL \url{https://doi.org/10.1109/TBME.2012.2205687}.

\bibitem[Odstrcilik et~al.(2013)Odstrcilik, Kolar, Budai, Hornegger, Jan,
  Gazarek, Kubena, Cernosek, Svoboda, and
  Angelopoulou]{odstrcilik_retinal_2013}
Jan Odstrcilik, Radim Kolar, Attila Budai, Joachim Hornegger, Jiri Jan, Jiri
  Gazarek, Tomas Kubena, Pavel Cernosek, Ondrej Svoboda, and Elli Angelopoulou.
\newblock Retinal vessel segmentation by improved matched filtering: evaluation
  on a new high‐resolution fundus image database.
\newblock \emph{IET Image Processing}, 7\penalty0 (4):\penalty0 373--383, June
  2013.
\newblock ISSN 1751-9667, 1751-9667.
\newblock \doi{10.1049/iet-ipr.2012.0455}.
\newblock URL
  \url{https://onlinelibrary.wiley.com/doi/10.1049/iet-ipr.2012.0455}.

\bibitem[Budai et~al.(2013)Budai, Bock, Maier, Hornegger, and
  Michelson]{budai_robust_2013}
A.~Budai, R.~Bock, A.~Maier, J.~Hornegger, and G.~Michelson.
\newblock Robust {Vessel} {Segmentation} in {Fundus} {Images}.
\newblock \emph{International Journal of Biomedical Imaging}, 2013:\penalty0
  1--11, 2013.
\newblock ISSN 1687-4188, 1687-4196.
\newblock \doi{10.1155/2013/154860}.
\newblock URL \url{http://www.hindawi.com/journals/ijbi/2013/154860/}.

\bibitem[Jin et~al.(2022)Jin, Huang, Zhou, Li, Yan, Sun, Zhang, Wang, and
  Ye]{jin_fives:_2022}
Kai Jin, Xingru Huang, Jingxing Zhou, Yunxiang Li, Yan Yan, Yibao Sun, Qianni
  Zhang, Yaqi Wang, and Juan Ye.
\newblock {FIVES}: {A} {Fundus} {Image} {Dataset} for {Artificial}
  {Intelligence} based {Vessel} {Segmentation}.
\newblock \emph{Scientific Data}, 9\penalty0 (1):\penalty0 475, August 2022.
\newblock ISSN 2052-4463.
\newblock \doi{10.1038/s41597-022-01564-3}.
\newblock URL \url{https://www.nature.com/articles/s41597-022-01564-3}.

\bibitem[Hoover et~al.(2000)Hoover, Kouznetsova, and
  Goldbaum]{hoover_locating_2000}
A.D. Hoover, V.~Kouznetsova, and M.~Goldbaum.
\newblock Locating blood vessels in retinal images by piecewise threshold
  probing of a matched filter response.
\newblock \emph{IEEE Transactions on Medical Imaging}, 19\penalty0
  (3):\penalty0 203--210, March 2000.
\newblock ISSN 02780062.
\newblock \doi{10.1109/42.845178}.
\newblock URL \url{http://ieeexplore.ieee.org/document/845178/}.

\bibitem[noa({\natexlab{e}})]{noauthor_cataract_nodate}
cataract dataset, {\natexlab{e}}.
\newblock URL \url{https://www.kaggle.com/datasets/jr2ngb/cataractdataset}.

\bibitem[Shukla and Tripathy()]{shukla_diabetic_2023}
Unnati~V. Shukla and Koushik Tripathy.
\newblock Diabetic retinopathy.
\newblock In \emph{{StatPearls}}. {StatPearls} Publishing.
\newblock URL \url{http://www.ncbi.nlm.nih.gov/books/NBK560805/}.

\bibitem[Yuen et~al.()Yuen, Pike, Khachikyan, and
  Nallasamy]{yuen_telehealth_2022}
Jenay Yuen, Sarah Pike, Steve Khachikyan, and Sudha Nallasamy.
\newblock Telehealth in ophthalmology.
\newblock In Simon~Lin Linwood, editor, \emph{Digital Health}. Exon
  Publications.
\newblock ISBN 978-0-645-33201-8.
\newblock URL \url{http://www.ncbi.nlm.nih.gov/books/NBK580631/}.

\bibitem[Müller et~al.()Müller, Wolf, Dolz-Marco, Tafreshi,
  Schmitz-Valckenberg, and Holz]{muller_ophthalmic_2019}
Philipp~L. Müller, Sebastian Wolf, Rosa Dolz-Marco, Ali Tafreshi, Steffen
  Schmitz-Valckenberg, and Frank~G. Holz.
\newblock Ophthalmic diagnostic imaging: Retina.
\newblock In Josef~F. Bille, editor, \emph{High Resolution Imaging in
  Microscopy and Ophthalmology: New Frontiers in Biomedical Optics}. Springer.
\newblock ISBN 978-3-030-16637-3 978-3-030-16638-0.
\newblock URL \url{http://www.ncbi.nlm.nih.gov/books/NBK554052/}.

\bibitem[Fang et~al.()Fang, Yokoi, Nagaoka, Shinohara, Onishi, Ishida, Yoshida,
  Xu, Jonas, and Ohno-Matsui]{fang_progression_2018}
Yuxin Fang, Tae Yokoi, Natsuko Nagaoka, Kosei Shinohara, Yuka Onishi, Tomoka
  Ishida, Takeshi Yoshida, Xian Xu, Jost~B. Jonas, and Kyoko Ohno-Matsui.
\newblock Progression of myopic maculopathy during 18-year follow-up.
\newblock \emph{Ophthalmology}, 125\penalty0 (6):\penalty0 863--877.
\newblock ISSN 0161-6420.
\newblock \doi{10.1016/j.ophtha.2017.12.005}.
\newblock URL
  \url{https://www.sciencedirect.com/science/article/pii/S0161642017329500}.

\bibitem[noa({\natexlab{f}})]{noauthor_pathologic_nodate}
Pathologic myopia - an overview {\textbackslash}textbar {ScienceDirect} topics,
  {\natexlab{f}}.
\newblock URL
  \url{https://www.sciencedirect.com/topics/medicine-and-dentistry/pathologic-myopia}.

\bibitem[Kolb()]{kolb_facts_1995}
Helga Kolb.
\newblock Facts and figures concerning the human retina.
\newblock In Helga Kolb, Eduardo Fernandez, and Ralph Nelson, editors,
  \emph{Webvision: The Organization of the Retina and Visual System}.
  University of Utah Health Sciences Center.
\newblock URL \url{http://www.ncbi.nlm.nih.gov/books/NBK11556/}.

\bibitem[Masland(2012)]{masland_neuronal_2012}
Richard~H. Masland.
\newblock The {Neuronal} {Organization} of the {Retina}.
\newblock \emph{Neuron}, 76\penalty0 (2):\penalty0 266--280, October 2012.
\newblock ISSN 08966273.
\newblock \doi{10.1016/j.neuron.2012.10.002}.
\newblock URL
  \url{https://linkinghub.elsevier.com/retrieve/pii/S0896627312008835}.

\bibitem[Ohno-Matsui et~al.(2015)Ohno-Matsui, Kawasaki, Jonas, Cheung, Saw,
  Verhoeven, Klaver, Moriyama, Shinohara, Kawasaki, Yamazaki, Meuer, Ishibashi,
  Yasuda, Yamashita, Sugano, Wang, Mitchell, Wong, and {META-analysis for
  Pathologic Myopia (META-PM) Study Group}]{pm1c}
Kyoko Ohno-Matsui, Ryo Kawasaki, Jost~B. Jonas, Chui Ming~Gemmy Cheung,
  Seang-Mei Saw, Virginie J.~M. Verhoeven, Caroline C.~W. Klaver, Muka
  Moriyama, Kosei Shinohara, Yumiko Kawasaki, Mai Yamazaki, Stacy Meuer,
  Tatsuro Ishibashi, Miho Yasuda, Hidetoshi Yamashita, Akira Sugano, Jie~Jin
  Wang, Paul Mitchell, Tien~Yin Wong, and {META-analysis for Pathologic Myopia
  (META-PM) Study Group}.
\newblock International photographic classification and grading system for
  myopic maculopathy.
\newblock \emph{American Journal of Ophthalmology}, 159\penalty0 (5):\penalty0
  877--883.e7, May 2015.
\newblock ISSN 1879-1891.
\newblock \doi{10.1016/j.ajo.2015.01.022}.

\bibitem[Flitcroft et~al.(2019)Flitcroft, He, Jonas, Jong, Naidoo, Ohno-Matsui,
  Rahi, Resnikoff, Vitale, and Yannuzzi]{pm22}
Daniel~Ian Flitcroft, Mingguang He, Jost~B. Jonas, Monica Jong, Kovin Naidoo,
  Kyoko Ohno-Matsui, Jugnoo Rahi, Serge Resnikoff, Susan Vitale, and Lawrence
  Yannuzzi.
\newblock {IMI} – {Defining} and {Classifying} {Myopia}: {A} {Proposed} {Set}
  of {Standards} for {Clinical} and {Epidemiologic} {Studies}.
\newblock \emph{Investigative Opthalmology \& Visual Science}, 60\penalty0
  (3):\penalty0 M20, February 2019.
\newblock ISSN 1552-5783.
\newblock \doi{10.1167/iovs.18-25957}.
\newblock URL
  \url{http://iovs.arvojournals.org/article.aspx?doi=10.1167/iovs.18-25957}.

\bibitem[Terasaki et~al.(2016)Terasaki, Yamashita, Yoshihara, Kii, Tanaka,
  Nakao, and Sakamoto]{pm73}
Hiroto Terasaki, Takehiro Yamashita, Naoya Yoshihara, Yuya Kii, Minoru Tanaka,
  Kumiko Nakao, and Taiji Sakamoto.
\newblock Location of {Tessellations} in {Ocular} {Fundus} and {Their}
  {Associations} with {Optic} {Disc} {Tilt}, {Optic} {Disc} {Area}, and {Axial}
  {Length} in {Young} {Healthy} {Eyes}.
\newblock \emph{PLOS ONE}, 11\penalty0 (6):\penalty0 e0156842, June 2016.
\newblock ISSN 1932-6203.
\newblock \doi{10.1371/journal.pone.0156842}.
\newblock URL \url{https://dx.plos.org/10.1371/journal.pone.0156842}.

\bibitem[noa({\natexlab{g}})]{noauthor_annotation_2021}
Annotation and quality control specifications for fundus color photograph.
\newblock \emph{Intelligent Medicine}, 1\penalty0 (2):\penalty0 80--87,
  {\natexlab{g}}.
\newblock ISSN 26671026.
\newblock \doi{10.1016/j.imed.2021.05.006}.
\newblock URL
  \url{https://linkinghub.elsevier.com/retrieve/pii/S2667102621000188}.

\bibitem[noa({\natexlab{h}})]{noauthor_jupyter_nodate}
Jupyter {Book}, {\natexlab{h}}.
\newblock URL \url{https://zenodo.org/records/4539666}.

\bibitem[Paszke et~al.(2019)Paszke, Gross, Massa, Lerer, Bradbury, Chanan,
  Killeen, Lin, Gimelshein, Antiga, Desmaison, Köpf, Yang, DeVito, Raison,
  Tejani, Chilamkurthy, Steiner, Fang, Bai, and Chintala]{paszke_pytorch:_2019}
Adam Paszke, Sam Gross, Francisco Massa, Adam Lerer, James Bradbury, Gregory
  Chanan, Trevor Killeen, Zeming Lin, Natalia Gimelshein, Luca Antiga, Alban
  Desmaison, Andreas Köpf, Edward Yang, Zach DeVito, Martin Raison, Alykhan
  Tejani, Sasank Chilamkurthy, Benoit Steiner, Lu~Fang, Junjie Bai, and Soumith
  Chintala.
\newblock {PyTorch}: {An} {Imperative} {Style}, {High}-{Performance} {Deep}
  {Learning} {Library}, December 2019.
\newblock URL \url{http://arxiv.org/abs/1912.01703}.
\newblock arXiv:1912.01703 [cs, stat].

\end{thebibliography}
